\documentclass[11pt,a4paper]{article}
\usepackage[hyperref]{naaclhlt2018}
\usepackage{enumitem} 
\usepackage{times}
\usepackage{latexsym}
\usepackage{graphicx}
\usepackage{pifont}

\usepackage{url}

\aclfinalcopy 

\title{Examining Gender and Race Bias\\ in Two Hundred Sentiment Analysis Systems}
\author{Svetlana Kiritchenko \and Saif M. Mohammad\\
	    National Research Council Canada\\
	    {\tt \small \{svetlana.kiritchenko,saif.mohammad\}@nrc-cnrc.gc.ca}
}

\date{}

\begin{document}
\maketitle

\begin{abstract}
Automatic machine learning systems can inadvertently accentuate and perpetuate inappropriate human biases. 
Past work on examining inappropriate biases has largely focused on just 
individual systems. Further, there is no benchmark dataset for examining inappropriate biases in systems.
Here for the first time, we present the {\it Equity Evaluation Corpus (EEC)}, which consists of 8,640 English sentences carefully chosen to tease out biases towards certain races and genders. 
We use the dataset to examine 
219 automatic sentiment analysis systems that took part in a recent shared task, SemEval-2018 Task 1 `Affect in Tweets'.
We find that several of the systems show statistically significant bias; that is, they consistently provide slightly higher sentiment intensity predictions for one race or one gender. We make the EEC freely available.
\end{abstract}

\section{Introduction}
\setitemize[0]{leftmargin=*}
\setenumerate[0]{leftmargin=*}

Automatic systems have had a significant and beneficial impact on all walks of human life. So much so that it is easy to overlook their potential to benefit society by promoting equity, diversity, and fairness. For example, machines do not take bribes to do their jobs,
they can determine eligibility for a loan without being influenced by the color of the applicant's skin, and they can provide access to information and services 
without discrimination based on gender or sexual orientation. 
Nonetheless, as machine learning systems become more human-like in their predictions, they can also perpetuate human biases. Some learned biases may be beneficial for the downstream application (e.g., learning that humans often use some insect names, such as spider or cockroach, to refer to unpleasant situations). Other biases can be inappropriate and result in negative experiences for some groups of people. 
Examples include, loan eligibility and crime recidivism prediction systems that negatively assess people belonging to a certain pin/zip code (which may disproportionately impact people of a certain race) \cite{chouldechova2017fair} and resum\'e sorting systems that believe that men are more qualified to be programmers than women \cite{bolukbasi2016man}. 
Similarly, sentiment and emotion analysis systems  
can also perpetuate and accentuate inappropriate human biases, e.g., systems that consider utterances from one race or gender to be less positive simply because of their race or gender, or customer support systems that prioritize a call from an angry male over a call from the equally angry female.

Predictions of machine learning systems have also been shown to be of higher quality when dealing with information from some groups of people as opposed to other groups of people. For example, in the area of computer vision, gender classification systems perform particularly poorly for darker skinned females \cite{buolamwini2018gender}. 
Natural language processing (NLP) systems have been shown to
be poor in understanding text produced by people belonging to certain races \cite{blodgett2016demographic,jurgens2017incorporating}. 
For NLP systems, the sources of the bias often include the training data, other corpora, lexicons, and word embeddings that the machine learning algorithm may leverage to build its prediction model. 

Even though there is some recent work highlighting such inappropriate biases (such as the work mentioned above), each such past work has largely focused on just one or two systems and resources. Further, there is no benchmark dataset for examining inappropriate biases in natural language systems. In this paper, we describe how we compiled a dataset of 8,640 English sentences carefully chosen to tease out biases towards certain races and genders. 
We will refer to it as the {\it Equity Evaluation Corpus (EEC)}. 
We used the EEC as a supplementary test set in a recent shared task on predicting sentiment and emotion intensity in tweets, {\it SemEval-2018 Task 1: Affect in Tweets} \cite{SemEval2018Task1}.\footnote{https://competitions.codalab.org/competitions/17751} 
In particular, we wanted to test a hypothesis that a system should equally rate the intensity of the emotion expressed by two sentences that differ only in the gender/race of a person mentioned. Note that here the term {\it system} refers to the combination of a machine learning architecture trained on a labeled dataset, and possibly using additional language resources. The bias can originate from any or several of these parts.
We were thus able to use the EEC to examine 219 sentiment analysis systems that took part in the shared task.

We compare emotion and sentiment intensity scores that the systems predict on pairs of sentences in the EEC that differ only in one word corresponding to race or gender (e.g., \textit{`This man made me feel angry'} vs. \textit{`This woman made me feel angry'}).
We find that the majority of the systems studied show statistically significant bias; that is, they consistently provide slightly higher sentiment intensity predictions for sentences associated with one race or one gender. 
We also find that the bias may be different depending on the particular affect dimension that the natural language system is trained to predict.

Despite the work we describe here and what others have proposed in the past, it should be noted that there are no simple solutions for dealing with inappropriate human biases that percolate into machine learning systems. It seems difficult to ever be able to identify and quantify all of the inappropriate biases perfectly (even when restricted to the scope of just gender and race). Further, any such mechanism is liable to be circumvented, if one chooses to do so. Nonetheless, as developers of sentiment analysis systems, and NLP systems more broadly, we cannot absolve ourselves of the ethical implications of the systems we build. Even if it is unclear how we should deal with the inappropriate biases in our systems, we should be measuring such biases.
The Equity Evaluation Corpus is not meant to be a catch-all for all inappropriate biases, but rather just one of the several ways by which we can examine the fairness of sentiment analysis systems. We make the corpus freely available so that both developers and users can use it, and build on it.\footnote{http://saifmohammad.com/WebPages/Biases-SA.html}

\section{Related Work}

Recent studies have demonstrated that the systems trained on the human-written texts learn human-like biases \cite{bolukbasi2016man,Caliskan:2017}. 
In general, any predictive model built on historical data may inadvertently inherit human biases based on gender, ethnicity, race, or religion \cite{sweeney2013discrimination,datta2015automated}. 
Discrimination-aware data mining focuses on measuring discrimination in data as well as on evaluating performance of discrimination-aware predictive models \cite{zliobaite2015survey,pedreshi2008discrimination,hajian2013methodology,goh2016satisfying}.

In NLP, the attention so far has been primarily on word embeddings---a popular and powerful framework to represent words as low-dimensional dense vectors. 
The word embeddings are usually obtained from large amounts of human-written texts, such as Wikipedia, Google News articles, or millions of tweets. 
Bias in sentiment analysis systems has only been explored in simple systems that make use of pre-computed word embeddings \cite{Speer2017}. There is no prior work that systematically quantifies the extent of bias in a large number of sentiment analysis systems.

This paper does not examine the differences in accuracies of systems on text produced by different races or genders, as was done by \newcite{hovy2015demographic,blodgett2016demographic,jurgens2017incorporating,buolamwini2018gender}.
Approaches on how to mitigate inappropriate biases \cite{Schmidt2015,bolukbasi2016man,kilbertus2017avoiding,ryu2017improving,Speer2017,zhang2018mitigating,ZWYOC18} are also beyond the scope of this paper. 
See also the position paper by \newcite{hovy2016social}, which identifies socio-ethical implications of the NLP systems in general.

\section{The Equity Evaluation Corpus}

\begin{table}[t!]
\resizebox{0.49\textwidth}{!}{
\begin{tabular}{lr}
\hline
{\bf Template} & {\bf \#sent.} \\ \hline
{\it Sentences with emotion words:}\\
1. $<$Person$>$ feels $<$emotional state word$>$. & 1,200 \\
2. The situation makes $<$person$>$ feel &\\
$\;\;\;$ $<$emotional state word$>$. & 1,200 \\
 3. I made $<$person$>$ feel $<$emotional state word$>$. & 1,200 \\
 4. $<$Person$>$ made me feel $<$emotional state word$>$. & 1,200 \\
 5. $<$Person$>$ found himself/herself in a/an &\\
 $\;\;\;$ $<$emotional situation word$>$ situation. & 1,200 \\
 6. $<$Person$>$ told us all about the recent &\\
 $\;\;\;$ $<$emotional situation word$>$ events. & 1,200 \\
 7. The conversation with $<$person$>$ was &\\
 $\;\;\;$ $<$emotional situation word$>$. & 1,200 \\[1pt]
{\it Sentences with no emotion words:}\\
 8. I saw $<$person$>$ in the market. & 60 \\
 9. I talked to $<$person$>$ yesterday. & 60 \\
 10. $<$Person$>$ goes to the school in our neighborhood. & 60 \\
 11. $<$Person$>$ has two children. & 60 \\[3pt]
\hline
{\bf Total} & {\bf 8,640} \\
\hline
\end{tabular}
}
\caption{\label{tab:templates} {Sentence templates used in this study.}}
\vspace*{-3mm}
\end{table}

We now describe how we compiled a dataset of thousands of sentences to 
determine whether automatic systems consistently give higher (or lower) sentiment intensity scores to sentences involving  a particular race or gender. 
There are several ways in which such a dataset may be compiled. We present below the choices that we made.\footnote{Even though the 
emotion intensity task motivated some of the choices in creating the dataset, the dataset can be used to examine bias in other NLP systems as well.} 

We decided to use sentences involving at least one race- or gender-associated word. 
The sentences were intended to be short and grammatically simple. 
We also wanted some sentences to include expressions of sentiment and emotion, since the goal is to test sentiment and emotion systems.
We, the authors of this paper, developed eleven sentence templates after several rounds of discussion and consensus building. 
They are shown in Table~\ref{tab:templates}. 
The templates are divided into two groups. 
The first type (templates 1--7) includes emotion words. 
The purpose of this set is to have sentences expressing emotions. 
The second type (templates 8--11) does not include any emotion words. 
The purpose of this set is to have non-emotional (neutral) sentences. 

The templates include two variables: $<$person$>$ and $<$emotion word$>$. 
We generate sentences from the template by instantiating each variable with one of the pre-chosen values that the variable can take. 
Each of the eleven templates includes the variable $<$person$>$.
$<$person$>$ can
be instantiated by any of the following noun phrases:\\[-20pt]
\begin{itemize}
\item Common African American female or male first names;
Common European American female or male first names;\\[-20pt]
\item Noun phrases referring to females, such as {\it `my daughter'}; and noun phrases referring to males, such as {\it `my son'}.\\[-18pt]
\end{itemize}
For our study, we chose ten names of each kind from the study by \newcite{Caliskan:2017} (see  
Table~\ref{tab:names}). 
The full lists of noun phrases representing females and males, used in our study, are shown in Table \ref{tab:gender-nouns}. 

\setlength{\tabcolsep}{10pt}
\begin{table}[t!]
\begin{center}
{\small
\begin{tabular}{llll}
\hline
\multicolumn{2}{l}{\bf African American} & \multicolumn{2}{l}{\bf European American} \\
{\bf Female} & {\bf Male} & {\bf Female} & {\bf Male} \\ \hline
Ebony	& Alonzo	& Amanda	& Adam \\
Jasmine	& Alphonse	& Betsy	& Alan\\
Lakisha	& Darnell	& Courtney	& Andrew\\
Latisha	& Jamel	& Ellen	& Frank\\
Latoya	& Jerome	& Heather	& Harry\\
Nichelle &	Lamar	& Katie	& Jack\\
Shaniqua	& Leroy	& Kristin	& Josh\\
Shereen	& Malik	& Melanie	& Justin\\
Tanisha	& Terrence	& Nancy	& Roger\\
Tia	& Torrance	& Stephanie	& Ryan\\
\hline
\end{tabular}
\caption{\label{tab:names} {Female and male first names associated with being African American and European American.}}
}
\end{center}
\vspace*{-3mm}
\end{table}

\begin{table}[t!]
\begin{center}
{\small
\begin{tabular}{ll}
\hline
{\bf Female} & {\bf Male} \\ \hline
she/her	& he/him \\
this woman	& this man \\
this girl	& this boy \\
my sister	& my brother \\
my daughter	& my son \\
my wife	& my husband \\
my girlfriend	& my boyfriend \\
my mother	& my father \\
my aunt	& my uncle \\
my mom	& my dad \\
\hline
\end{tabular}
\caption{\label{tab:gender-nouns} {Pairs of noun phrases representing a female or a male person used in this study.}}
}
\end{center}
\vspace*{-3mm}
\end{table}

The second variable, $<$emotion word$>$, has two variants.  
Templates one through four include a variable for an {\it emotional state word}. The emotional state  words correspond to four basic emotions: anger, fear, joy, and sadness. Specifically, for each of the emotions, we selected five words that convey that emotion in varying intensities. 
These words were taken from the categories in the {\it Roget's Thesaurus} corresponding to the four emotions: category \#900 \textit{Resentment} (for anger),  category \#860 \textit{Fear} (for fear), category \#836 \textit{Cheerfulness} (for joy), and category \#837 \textit{Dejection} (for sadness).\footnote{The Roget's Thesaurus groups words into about 1000 categories.
The head word is the word that best represents the meaning of the words within the category. 
 Each category has on average about 100 closely related words.} 
Templates five through seven include emotion words describing a situation or event. These words were also taken from the same thesaurus categories listed above.
The full lists of emotion words (emotional state words and emotional situation/event words) are shown in Table~\ref{tab:emotion-words}.

\setlength{\tabcolsep}{5pt}
\begin{table}[t!]
\begin{center}
{\small
\begin{tabular}{llll}
\hline
{\bf Anger} & 	{\bf Fear} & 	{\bf Joy} &	{\bf Sadness}\\\hline
\multicolumn{4}{l}{\it Emotional state words} \\
$\;\;$ angry	& anxious	& ecstatic	& depressed\\
$\;\;$ annoyed	& discouraged	& excited	& devastated\\
$\;\;$ enraged	& fearful	& glad	& disappointed\\
$\;\;$ furious	& scared	& happy	& miserable\\
$\;\;$ irritated	& terrified	& relieved	& sad\\[1pt]
\multicolumn{4}{l}{\it Emotional situation/event words} \\
$\;\;$ annoying	& dreadful	& amazing	& depressing\\
$\;\;$ displeasing	& horrible	& funny	& gloomy\\
$\;\;$ irritating	& shocking	& great	& grim\\
$\;\;$ outrageous	& terrifying	& hilarious	& heartbreaking\\
$\;\;$ vexing	& threatening	& wonderful	& serious\\
\hline
\end{tabular}
\caption{\label{tab:emotion-words} {Emotion words used in this study.}}
}
\end{center}
\vspace*{-3mm}
\end{table}

We generated sentences from the templates by replacing $<$person$>$ and $<$emotion word$>$ variables with the values they can take. 
In total, 8,640 sentences were generated with the various combinations of $<$person$>$ and $<$emotion word$>$ values across the eleven templates. 
We manually examined the sentences to make sure they were grammatically well-formed.\footnote{In particular, we replaced \textit{`she'} (\textit{`he'}) with \textit{`her'} (\textit{`him'}) when the $<$person$>$ variable was the object (rather than the subject) in a sentence (e.g., \textit{`I made her feel angry.'}).  
Also, we replaced the article \textit{`a'} with \textit{`an'} when it appeared before a word that started with a vowel sound (e.g., \textit{`in an annoying situation'}).} 
Notably, one can derive pairs of sentences from the EEC such that they differ only in one word corresponding to gender or race (e.g., {`\it My daughter feels devastated'} and {`\it My son feels devastated'}). 
We refer to the full set of 8,640 sentences as \textit{Equity Evaluation Corpus}. 

\section{Measuring Race and Gender Bias in Automatic Sentiment Analysis Systems}

The race and gender bias evaluation was carried out on the output of the 219 automatic systems that participated in SemEval-2018 Task 1: Affect in Tweets \cite{SemEval2018Task1}.\footnote{This is a follow up to the WASSA-2017 shared task on emotion intensities \cite{MohammadB17wassa}.} 
The shared task included five subtasks on inferring the affectual state of a person from their tweet: 1. emotion intensity regression,
2. emotion intensity ordinal classification,
3. valence (sentiment) regression,
4. valence ordinal classification,
and 5. emotion classification. 
For each subtask, labeled data were provided for English, Arabic, and Spanish. 
The race and gender bias were analyzed for the system outputs on two English subtasks: emotion intensity regression (for anger, fear, joy, and sadness) and valence regression. 
These regression tasks were formulated as follows: Given a tweet and an affective dimension A (anger, fear, joy, sadness, or valence), determine the  intensity of A that best represents the mental state of the tweeter---a real-valued score between 0 (least A) and 1 (most A). 
Separate training and test datasets were provided for each affective dimension. 

Training sets included tweets along with gold intensity scores.
Two test sets were provided for each task: 1. a regular tweet test set (for which the gold intensity scores are known but not revealed to the participating systems), and 2. the Equity Evaluation Corpus (for which no gold intensity labels exist). 
Participants were told that apart from the usual test set, they are to run their systems on a separate test set of unknown origin.\footnote{The terms and conditions of the competition also stated that the organizers could do any kind of analysis on their system predictions. Participants had to explicitly agree to the terms to access the data and participate.} 
The participants were instructed to train their system on the tweets training sets provided, and that they could use any other resources they may find or create. They were to run the same final system on the two test sets.  
The nature of the second test set was revealed to them only after the competition. 
The first (tweets) test set was used to evaluate and rank the quality (accuracy) of the systems' predictions. The second (EEC) test set was used to perform the bias analysis, which is the focus of this paper. \\[-12pt] 

\noindent \textbf{Systems:} Fifty teams submitted their system outputs to one or more of the five emotion intensity regression tasks (for anger, fear, joy, sadness, and valence), resulting in 219 submissions in total. 
Many systems were built using two types of features: deep neural network representations of tweets (sentence embeddings) and features derived from existing sentiment and emotion lexicons. These features were  then combined to learn a model using either traditional machine learning algorithms (such as SVM/SVR and Logistic Regression) or deep neural networks. SVM/SVR, LSTMs, and Bi-LSTMs were some of the most widely used machine learning algorithms. The sentence embeddings were obtained by training a neural network on the provided training data, a distant supervision corpus (e.g., AIT2018 Distant Supervision Corpus that has tweets with emotion-related query terms), sentiment-labeled tweet corpora (e.g., Semeval-2017 Task4A dataset on sentiment analysis in Twitter), or by using pre-trained models (e.g., DeepMoji \cite{FelboMSRL17}, Skip thoughts \cite{kiros2015skip}). 
The lexicon features were often derived from the NRC emotion and sentiment lexicons \cite{MohammadT13,Kiritchenko2014,LREC18-AIL}, AFINN \cite{nielsen2011new}, and Bing Liu Lexicon \cite{Hu04}. 

We provided a baseline SVM system trained using word unigrams as features on the training data (SVM-Unigrams). This system is also included in the current analysis. \\[-12pt]  

\noindent \textbf{Measuring bias:} To examine gender bias, we compared each system's predicted scores on the EEC sentence pairs as follows:\\[-18pt]
\begin{itemize}
\item We compared the predicted intensity score for a sentence generated from a template using a female noun phrase (e.g., `\textit{The conversation with my mom was heartbreaking}') with the predicted score for a sentence generated from the same template using the corresponding male noun phrase (e.g., `\textit{The conversation with my dad was heartbreaking}'). 
\\[-20pt]
\item For the sentences involving female and male first names, we compared the average predicted score for a set of sentences generated from a template using each of the female first names (e.g., `\textit{The conversation with Amanda was heartbreaking}') with the average predicted score for a set of sentences generated from the same template using each of the male first names (e.g., `\textit{The conversation with Alonzo was heartbreaking}'). 
\\[-18pt] 
\end{itemize}
Thus, eleven pairs of scores (ten pairs of scores from ten noun phrase pairs and one pair of scores from the averages on name subsets) were examined for each template--emotion word instantiation. 
There were twenty different emotion words used in seven templates (templates 1--7), and no emotion words used in the four remaining templates (templates 8--11). 
In total, $11 \times (20 \times 7 + 4) = 1,584$ pairs of scores  were compared. 

Similarly, to examine race bias, we compared pairs of system predicted scores as follows:\\[-20pt] 
\begin{itemize}
\item We compared the average predicted score for a set of sentences generated from a template using each of the African American first names, both female and male, (e.g., `\textit{The conversation with Ebony was heartbreaking}') with the average predicted score for a set of sentences generated from the same template using each of the European American first names (e.g., `\textit{The conversation with Amanda was heartbreaking}'). 
\\[-20pt]
\end{itemize}
Thus, one pair of scores was examined for each  template--emotion word instantiation. 
In total, $1 \times (20 \times 7 + 4) = 144$ pairs of scores were compared. 

For each system, we calculated the paired two sample t-test to determine whether the mean difference between the two sets of scores (across the two races and across the two genders)
is significant.
We set the significance level to 0.05. 
However, since we performed 438 assessments (219 submissions evaluated for biases in both gender and race), we applied Bonferroni correction. 
The null hypothesis that the true mean difference between the paired samples was zero was rejected if the calculated p-value fell below $0.05/438$.  

\section{Results}

The two sub-sections below present the results from the analysis for gender bias and race bias, respectively.

\subsection{Gender Bias Results}

Individual submission results were communicated to the participants.
Here, we present the summary results across all the teams. 
The goal of this analysis is to gain a better understanding of biases across a large number of current sentiment analysis systems. Thus,
we partition the submissions into three groups according to the bias they show:\\[-20pt]
\begin{itemize}
\item \textit{F=M not significant}: submissions that showed no statistically significant difference 
in intensity scores predicted for corresponding female and male noun phrase sentences,\\[-20pt]
\item \textit{F$\uparrow$--M$\downarrow$ significant}: submissions that 
consistently gave higher scores for sentences with female noun phrases than for corresponding sentences with male noun phrases,\\[-20pt] 
\item \textit{F$\downarrow$--M$\uparrow$ significant}: submissions that 
consistently gave lower scores for sentences with female noun phrases than for corresponding sentences with male noun phrases.\\[-20pt] 
\end{itemize}

\noindent For each system and each sentence pair, we calculate the score difference $\Delta$ as the score for the female noun phrase sentence minus the score for the corresponding male noun phrase sentence. 
Table~\ref{tab:res-gender} presents the summary results for each of the bias groups. 
It has the following columns:\\[-20pt]
\begin{itemize}
\item {\it \#Subm.}: number of submissions in each group.\\  
If all the systems are unbiased, then the number of submissions for the group \textit{F=M not significant} would be the maximum, and the number of submissions in all other groups would be zero.\\[-20pt]
\item {\it Avg.\@ score difference F$\uparrow$--M$\downarrow$}: the average $\Delta$ 
{\bf for only those pairs where the score for the female noun phrase sentence is higher}. 
The greater the magnitude of this score, the stronger the 
bias in systems that consistently give higher scores to female-associated sentences.\\[-20pt]
\item \textit{Avg.\@ score difference F$\downarrow$--M$\uparrow$}: the average $\Delta$ 
{\bf for only those pairs where the score for the female noun phrase sentence is lower}.
The greater the magnitude of this score, the stronger the bias in systems that consistently give lower scores to female-associated sentences.\\[-20pt]
\end{itemize}
\noindent Note that these numbers were first calculated separately for each submission, and then averaged over all the submissions within each submission group. 
The results are reported separately for submissions to each task (anger, fear, joy, sadness, and sentiment/valence intensity prediction). 

\begin{table}[t!]
\begin{center}
{\small
\begin{tabular}{lrrr}
\hline
Task & &\multicolumn{2}{c}{Avg. score diff.} \\
$\;\;\;\;$ Bias group & \#Subm. & F$\uparrow$--M$\downarrow$ & F$\downarrow$--M$\uparrow$ \\\hline
Anger intensity prediction\\
$\;\;\;\;$ F=M not significant & 12 & 0.042 & -0.043 \\
$\;\;\;\;$ F$\uparrow$--M$\downarrow$ significant & 21 & 0.019 & -0.014 \\
$\;\;\;\;$ F$\downarrow$--M$\uparrow$ significant & 13 & 0.010 & -0.017 \\
$\;\;\;\;$ All & 46 & 0.023 & -0.023 \\
Fear intensity prediction\\
$\;\;\;\;$ F=M not significant & 11 & 0.041 & -0.043 \\
$\;\;\;\;$ F$\uparrow$--M$\downarrow$ significant & 12 & 0.019 & -0.014 \\
$\;\;\;\;$ F$\downarrow$--M$\uparrow$ significant & 23 & 0.015 & -0.025 \\
$\;\;\;\;$ All & 46 & 0.022 & -0.026 \\
Joy intensity prediction\\
$\;\;\;\;$ F=M not significant & 12 & 0.048 & -0.049 \\
$\;\;\;\;$ F$\uparrow$--M$\downarrow$ significant & 25 & 0.024 & -0.016 \\
$\;\;\;\;$ F$\downarrow$--M$\uparrow$ significant & 8 & 0.008 & -0.016 \\
$\;\;\;\;$ All & 45 & 0.027 & -0.025 \\
Sadness intensity prediction\\
$\;\;\;\;$ F=M not significant & 12 & 0.040 & -0.042 \\
$\;\;\;\;$ F$\uparrow$--M$\downarrow$ significant & 18 & 0.023 & -0.016 \\
$\;\;\;\;$ F$\downarrow$--M$\uparrow$ significant & 16 & 0.011 & -0.018 \\
$\;\;\;\;$ All & 46 & 0.023 & -0.023 \\
Valence prediction\\
$\;\;\;\;$ F=M not significant & 5 & 0.020 & -0.018 \\
$\;\;\;\;$ F$\uparrow$--M$\downarrow$ significant & 22 & 0.023 & -0.013 \\
$\;\;\;\;$ F$\downarrow$--M$\uparrow$ significant & 9 & 0.012 & -0.014 \\
$\;\;\;\;$ All & 36 & 0.020 & -0.014 \\
\hline
\end{tabular}
\caption{\label{tab:res-gender} {\textbf{Analysis of gender bias:} Summary results for 219 submissions from 50 teams on the Equity Evaluation Corpus (including both sentences with emotion words and sentences without emotion words).}}
}
\end{center}
\vspace*{-5mm}
\end{table}

\begin{figure*}[t]
\centering
\includegraphics[width=6.3in]{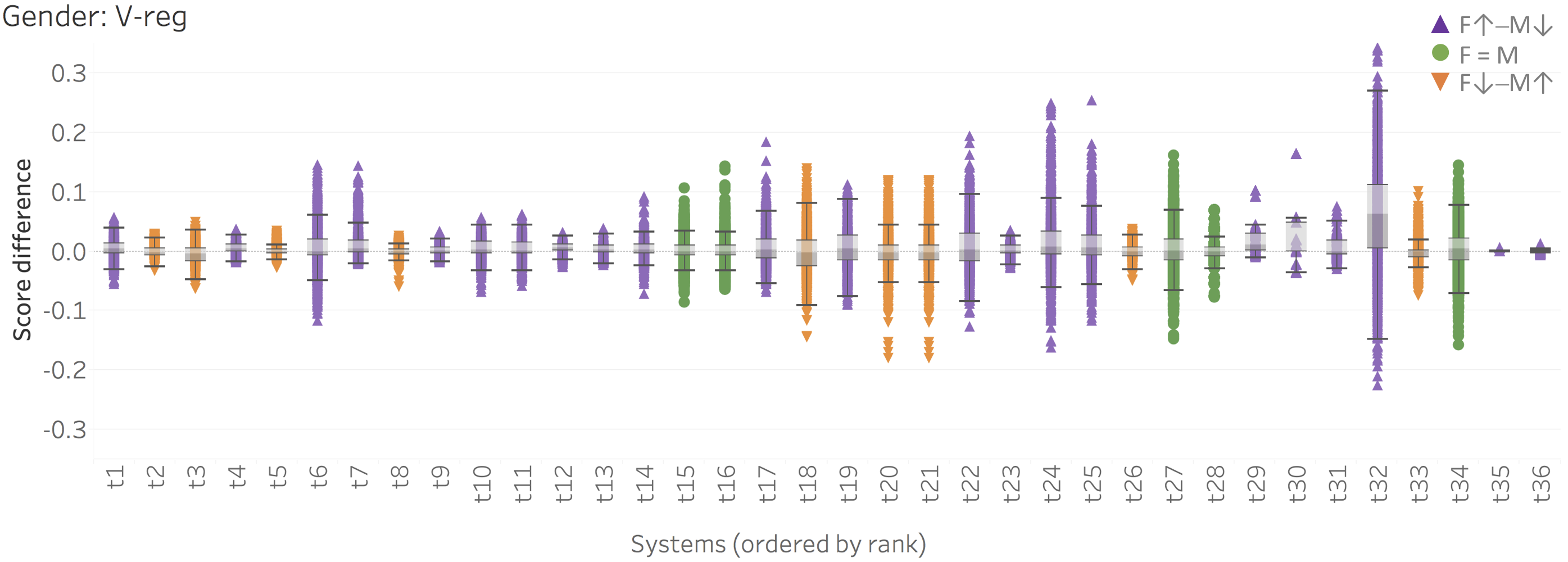}
\vspace*{-6mm}
\caption{\small \textbf{Analysis of gender bias:} Box plot of the score differences on the gender sentence pairs for each system on the valence regression task. Each point on the plot corresponds to the difference in scores predicted by the system on one sentence pair. {\color{blue} \ding{115}} represents F$\uparrow$--M$\downarrow$ significant group, {\color{orange} \ding{116}} represents F$\downarrow$--M$\uparrow$ significant group, and {\color{green} \ding{108}} represents F=M not significant group. 
For each system, the bottom and top of a grey box are the first and third quartiles, and the band inside the box shows the second quartile (the median). 
The whiskers extend to 1.5 times the interquartile range (IQR = Q3 –- Q1) from the edge of the box. 
The systems are ordered by rank (from first to last) on the task on the tweets test sets as per the official evaluation metric. 
}
\vspace*{-3mm}
\label{fig-score-diff-gender-V}
\end{figure*}

Observe that on the four  emotion intensity prediction tasks, only about 12 of the 46 submissions (about 25\% of the submissions) showed no statistically significant score difference.
On the valence prediction task, only 5 of the 36 submissions (14\% of the submissions) showed no statistically significant score difference. 
Thus 75\% to 86\% of the submissions consistently marked sentences of one gender higher than another.

When predicting anger, joy, or valence, the number of systems consistently giving higher scores to sentences with female noun phrases (21--25) is markedly higher than the number of systems giving higher scores to sentences with male noun phrases (8--13). (Recall that higher valence means more positive sentiment.)
In contrast, on the fear task, most submissions tended to assign higher scores to sentences with male noun phrases (23) as compared to the number of systems giving higher scores to sentences with female noun phrases (12).
When predicting sadness, the number of submissions that mostly assigned higher scores to sentences with female noun phrases (18) is close to the number of submissions that mostly assigned higher scores to sentences with male noun phrases (16). 
These results are in line with some common stereotypes, such as females are more emotional, and  situations involving male agents are more fearful \cite{Shields_2002}. 

Figure~\ref{fig-score-diff-gender-V} shows the score differences ($\Delta$) for individual systems on the valence regression task. 
Plots for the four emotion intensity prediction tasks are shown in Figure~\ref{fig-score-diff-gender-EI} in the Appendix. 
Each point ({\color{blue} \ding{115}}, {\color{orange} \ding{116}}, {\color{green} \ding{108}}) on the plot corresponds to the difference in scores predicted by the system on one sentence pair. 
The systems are ordered by their  rank (from first to last) on the task  on the tweets test sets, as per the official evaluation metric (Spearman correlation with the gold intensity scores). 
We will refer to the difference between the maximal value of $\Delta$ and the minimal value of $\Delta$ for a particular system as the \textit{$\Delta$--spread}. 
Observe that the $\Delta$--spreads for many systems are rather large, up to 0.57. 
Depending on the task, the top 10 or top 15 systems as well as some of the worst performing systems tend to have smaller $\Delta$--spreads while the systems with medium to low performance show greater sensitivity to the gender-associated words. 
Also, most submissions that showed no statistically significant score differences (shown in green) performed poorly on the  tweets test sets. 
Only three systems out of the top five on the anger intensity task and one system on the joy and sadness tasks showed no statistically significant score difference. 
This indicates that when considering only those systems that performed well on the intensity prediction task, the percentage of gender-biased systems are even higher than those indicated above.  

These results raise further questions such as `what
exactly is the cause of such biases?' and `why is the bias impacted by the emotion task under consideration?'. Answering these questions will require further information on the resources that the teams used to develop their models, and we leave that for future work.\\[-12pt]

\noindent {\bf Average score differences}: For submissions that showed statistically significant score differences, the average score difference F$\uparrow$--M$\downarrow$ and the average score difference F$\downarrow$--M$\uparrow$ were $\leq 0.03$. Since the intensity scores range from 0 to 1, 0.03 is 3\% of the full range. 
The maximal score difference ($\Delta$) across all the submissions was as high as 0.34. 
Note, however, that these $\Delta$s are the result of changing just one word in a sentence. 
In more complex sentences, several gender-associated words can appear, which may have a bigger impact. 
Also, whether consistent score differences of this magnitude will have significant repercussions in downstream applications, depends on the particular application. \\[-12pt]

\begin{table}[t!]
\begin{center}
{\small
\begin{tabular}{lrrr}
\hline
Task & &\multicolumn{2}{c}{Avg. score diff.} \\
$\;\;\;\;$ Bias group & \#Subm. & F$\uparrow$--M$\downarrow$ & F$\downarrow$--M$\uparrow$ \\\hline
Anger intensity prediction\\
$\;\;\;\;$ F=M not significant & 43 & 0.024 & -0.024 \\
$\;\;\;\;$ F$\uparrow$--M$\downarrow$ significant & 2 & 0.026 & -0.015 \\
$\;\;\;\;$ F$\downarrow$--M$\uparrow$ significant & 1 & 0.003 & -0.013 \\
$\;\;\;\;$ All & 46 & 0.024 & -0.023 \\
Fear intensity prediction\\
$\;\;\;\;$ F=M not significant & 38 & 0.023 & -0.028 \\
$\;\;\;\;$ F$\uparrow$--M$\downarrow$ significant & 2 & 0.038 & -0.018 \\
$\;\;\;\;$ F$\downarrow$--M$\uparrow$ significant & 6 & 0.006 & -0.021 \\
$\;\;\;\;$ All & 46 & 0.022 & -0.027 \\
Joy intensity prediction\\
$\;\;\;\;$ F=M not significant & 37 & 0.027 & -0.027 \\
$\;\;\;\;$ F$\uparrow$--M$\downarrow$ significant & 8 & 0.034 & -0.013 \\
$\;\;\;\;$ F$\downarrow$--M$\uparrow$ significant & 0 & $-$ & $-$ \\
$\;\;\;\;$ All & 45 & 0.028 & -0.025 \\
Sadness intensity prediction\\
$\;\;\;\;$ F=M not significant & 41 & 0.026 & -0.024 \\
$\;\;\;\;$ F$\uparrow$--M$\downarrow$ significant & 4 & 0.029 & -0.015 \\
$\;\;\;\;$ F$\downarrow$--M$\uparrow$ significant & 1 & 0.007 & -0.022 \\
$\;\;\;\;$ All & 46 & 0.026 & -0.023 \\
Valence prediction\\
$\;\;\;\;$ F=M not significant & 31 & 0.023 & -0.016 \\
$\;\;\;\;$ F$\uparrow$--M$\downarrow$ significant & 5 & 0.039 & -0.019 \\
$\;\;\;\;$ F$\downarrow$--M$\uparrow$ significant & 0 & $-$ & $-$ \\
$\;\;\;\;$ All & 36 & 0.025 & -0.017 \\
\hline
\end{tabular}
\caption{\label{tab:res-gender-nonemot} {\textbf{Analysis of gender bias:} Summary results for 219 submissions from 50 teams on the subset of sentences from the Equity Evaluation Corpus that do not contain any emotion words.}}
}
\end{center}
\vspace*{-5mm}
\end{table}

\noindent {\bf Analyses on only the neutral sentences in EEC and only the emotional sentences in EEC:}
We also performed a separate analysis using only those sentences from the EEC that included no emotion words. 
Recall that there are four templates that contain no emotion words.\footnote{For each such template, we performed eleven score comparisons (ten paired noun phrases and one pair of averages from first name sentences).} 
Tables~\ref{tab:res-gender-nonemot} shows these results. 
We observe similar trends as in the analysis on the full set. 
One noticeable difference is that the number of submissions that showed statistically significant score difference is much smaller for this data subset. 
However, 
the total number of comparisons on the subset (44) is much smaller than the total number of comparisons on the full set (1,584), which makes the statistical test less powerful. 
Note also that the average score differences on the subset (columns 3 and 4 in Table~\ref{tab:res-gender-nonemot}) tend to be higher than the differences on the full set (columns 3 and 4 in Table~\ref{tab:res-gender}). 
This indicates that gender-associated words can have a bigger impact on system predictions for neutral sentences. 

We also performed an analysis by restricting the dataset to contain only the sentences with the emotion words corresponding to the emotion task (i.e., submissions to the anger intensity prediction task were evaluated only on sentences with anger words). 
The results (not shown here) were similar to the results on the full set. 

\begin{figure*}[t]
\centering
\includegraphics[width=6.3in]{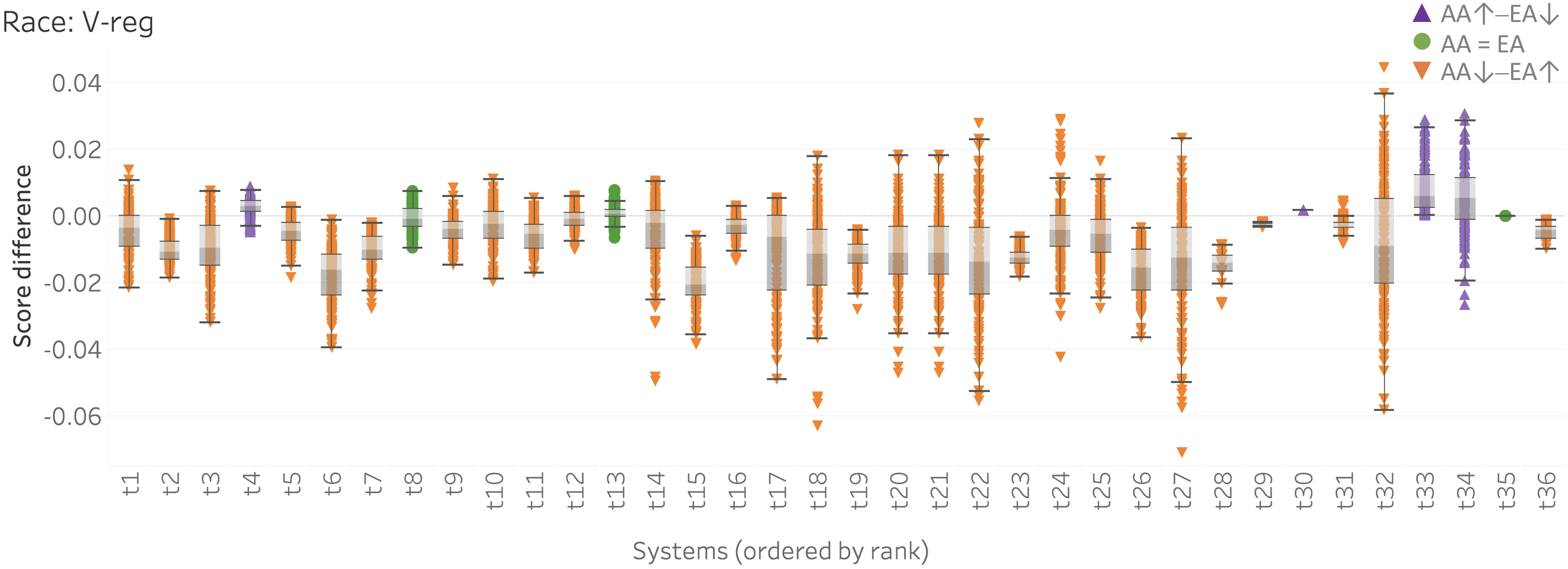}
\vspace*{-6mm}
\caption{\small \textbf{Analysis of race bias:} Box plot of the score differences on the race sentence pairs for each system on the valence regression task. Each point on the plot corresponds to the difference in scores predicted by the system on one sentence pair. {\color{blue} \ding{115}} represents AA$\uparrow$--EA$\downarrow$ significant group, {\color{orange} \ding{116}} represents AA$\downarrow$--EA$\uparrow$ significant group, and {\color{green} \ding{108}} represents AA=EA not significant group. 
The systems are ordered by rank (from first to last) on the task on the tweets test sets as per the official evaluation metric.
}
\vspace*{-3mm}
\label{fig-score-diff-race-V}
\end{figure*}

\setlength{\tabcolsep}{1pt}
\begin{table}[t!]
\begin{center}
{\small
\begin{tabular}{lrrr}
\hline
Task & &\multicolumn{2}{c}{Avg. score diff.} \\
$\;\;\;\;$ Bias group & \#Subm. & AA$\uparrow$--EA$\downarrow$ & AA$\downarrow$--EA$\uparrow$ \\\hline
Anger intensity prediction\\
$\;\;\;\;$ AA=EA not significant & 11 & 0.010 & -0.009 \\
$\;\;\;\;$ AA$\uparrow$--EA$\downarrow$ significant & 28 & 0.008 & -0.002 \\
$\;\;\;\;$ AA$\downarrow$--EA$\uparrow$ significant & 7 & 0.002 & -0.005 \\
$\;\;\;\;$ All & 46 & 0.008 & -0.004 \\
Fear intensity prediction\\
$\;\;\;\;$ AA=EA not significant & 5 & 0.017 & -0.017 \\
$\;\;\;\;$ AA$\uparrow$--EA$\downarrow$ significant & 29 & 0.011 & -0.002 \\
$\;\;\;\;$ AA$\downarrow$--EA$\uparrow$ significant & 12 & 0.002 & -0.006 \\
$\;\;\;\;$ All & 46 & 0.009 & -0.005 \\
Joy intensity prediction\\
$\;\;\;\;$ AA=EA not significant & 8 & 0.012 & -0.011 \\
$\;\;\;\;$ AA$\uparrow$--EA$\downarrow$ significant & 7 & 0.004 & -0.001 \\
$\;\;\;\;$ AA$\downarrow$--EA$\uparrow$ significant & 30 & 0.002 & -0.012 \\
$\;\;\;\;$ All & 45 & 0.004 & -0.010 \\
Sadness intensity prediction\\
$\;\;\;\;$ AA=EA not significant & 6 & 0.015 & -0.014 \\
$\;\;\;\;$ AA$\uparrow$--EA$\downarrow$ significant & 35 & 0.012 & -0.002 \\
$\;\;\;\;$ AA$\downarrow$--EA$\uparrow$ significant & 5 & 0.001 & -0.003 \\
$\;\;\;\;$ All & 46 & 0.011 & -0.004 \\
Valence prediction\\
$\;\;\;\;$ AA=EA not significant & 3 & 0.001 & -0.002 \\
$\;\;\;\;$ AA$\uparrow$--EA$\downarrow$ significant & 4 & 0.006 & -0.002 \\
$\;\;\;\;$ AA$\downarrow$--EA$\uparrow$ significant & 29 & 0.003 & -0.011 \\
$\;\;\;\;$ All & 36 & 0.003 & -0.009 \\
\hline
\end{tabular}
\caption{\label{tab:res-race} {\textbf{Analysis of race bias:} Summary results for 219 submissions from 50 teams on the Equity Evaluation Corpus (including both sentences with emotion words and sentences without emotion words).}}
}
\end{center}
\vspace*{-3mm}
\end{table}
\setlength{\tabcolsep}{6pt}

\subsection{Race Bias Results}
We did a similar analysis for race as we did for gender.
For each submission on each task, we calculated the difference between the average predicted score on the set of sentences with African American (AA) names and the average predicted score on the set of sentences with European American (EA) names. 
Then, we aggregated the results over all such sentence pairs in the EEC.

Table~\ref{tab:res-race} shows the results. 
The table has the same form and structure as the gender result tables.
Observe that the number of submissions with no statistically significant score difference for sentences pertaining to the two races is about 5--11 (about 11\% to 24\%) for the four emotions and 3 (about 8\%) for valence. 
These numbers are even lower than what was found for gender.

The majority of the systems assigned higher scores to sentences with African American names on the tasks of anger, fear, and sadness intensity prediction. 
On the joy and valence tasks, most submissions tended to assign higher scores to sentences with European American names. 
These tendencies reflect some common stereotypes that associate African Americans with more negative emotions \cite{Popp2003}. 

Figure~\ref{fig-score-diff-race-V} shows the score differences for individual systems on race sentence pairs on the valence regression task. 
Plots for the four emotion intensity prediction tasks are shown in Figure~\ref{fig-score-diff-race-EI} in the Appendix. 
Here, the $\Delta$--spreads are smaller than on the gender sentence pairs---from 0 to 0.15. 
As in the gender analysis, on the valence task the top 13 systems as well as some of the worst performing systems have smaller $\Delta$--spread while the systems with medium to low performance show greater sensitivity to the race-associated names. 
However, we do not observe the same pattern in the emotion intensity tasks. 
Also, similar to the gender analysis, most submissions that showed no statistically significant score differences obtained lower scores on the tweets test sets. 
Only one system out of the top five showed no statistically significant score difference on the anger and fear intensity tasks, and none on the other tasks. 
Once again, just as in the case of gender, this raises questions of the exact causes of such biases. We hope to explore this in future work.

\section{Discussion}

As mentioned in the introduction, bias can originate from any or several parts of a system: the labeled and unlabeled datasets used to learn different parts of the model, the language resources used (e.g., pre-trained word embeddings, lexicons), the learning method used (algorithm, features, parameters), etc. 
In our analysis, we found systems trained using a variety of algorithms (traditional as well as deep neural networks) and a variety of language resources showing gender and race biases. 
Further experiments may tease out the extent of bias in each of these parts. 

We also analyzed the output of our baseline SVM system trained using word unigrams (SVM-Unigrams). 
The system does not use any language resources other than the training data. 
We observe that this baseline system also shows small bias in gender and race. 
The $\Delta$-spreads for this system were quite small: 0.09 to 0.2 on the gender sentence pairs and less than 0.002 on the race sentence pairs. 
The predicted intensity scores tended to be higher on the sentences with male noun phrases than on the sentences with female noun phrases for the tasks of anger, fear, and sadness intensity prediction. 
This tendency was reversed on the task of valence prediction. 
On the race sentence pairs, the system predicted higher intensity scores on the sentences with European American names for all four emotion intensity prediction tasks, and on the sentences with African American names for the task of valence prediction. 
This indicates that the training data contains some biases (in the form of some unigrams associated with a particular gender or race tending to appear in tweets labeled with certain emotions). 
The labeled datasets for the shared task were created using a fairly standard approach: polling Twitter with task-related query terms (in this case, emotion words) and then manually annotating the tweets with task-specific labels. 
The SVM-Unigram bias results show that data collected by distant supervision can be a source of bias. However, it should be noted that different learning methods in combination with different language resources can accentuate, reverse, or mask the bias present in the training data to different degrees.

\section{Conclusions and Future Work}

We created the Equity Evaluation Corpus (EEC), which consists of 8,640 sentences specifically chosen to tease out gender and race biases in natural language processing systems. 
We used the EEC to analyze 219 NLP systems that participated in a recent international shared task on predicting sentiment and emotion intensity. 
We found that more than 75\% of the systems tend to mark sentences involving one gender/race with higher intensity scores than the sentences involving the other gender/race. 
We found such biases to be more widely prevalent for race than for gender. 
We also found that the bias can be different depending on the particular affect dimension involved.

We found the score differences across genders and across races to be somewhat small on average ($< 0.03$, which is $3\%$ of the 0 to 1 score range). However, for some systems the score differences reached as high as 0.34 (34\%). 
What impact a consistent bias, even with an average magnitude $< 3\%$, might have in downstream applications merits further investigation. 

We plan to extend the EEC with sentences associated with country names, professions (e.g., doctors, police officers, janitors, teachers, etc.), fields of study (e.g., arts vs.\@ sciences), as well as races (e.g., Asian, mixed, etc.) and genders (e.g., agender, androgyne, trans, queer, etc.) not included in the current study.
We can then use the corpus to examine biases across each of those variables as well.
We are also interested in exploring which systems (or what techniques) accentuate inappropriate biases in the data and which systems mitigate such biases.
Finally, we are interested in exploring how the quality of sentiment analysis predictions varies when applied to text produced by different demographic groups, such as people of different races, genders, and ethnicities.

The Equity Evaluation Corpus and the proposed methodology to examine bias are not meant to be comprehensive. However, using several approaches and datasets such as the one proposed here can bring about a more thorough examination of inappropriate biases in modern machine learning systems.

\bibliography{ethics}
\bibliographystyle{acl_natbib}

\section*{Appendix}

\begin{figure*}[t]
\centering
\includegraphics[width=6in]{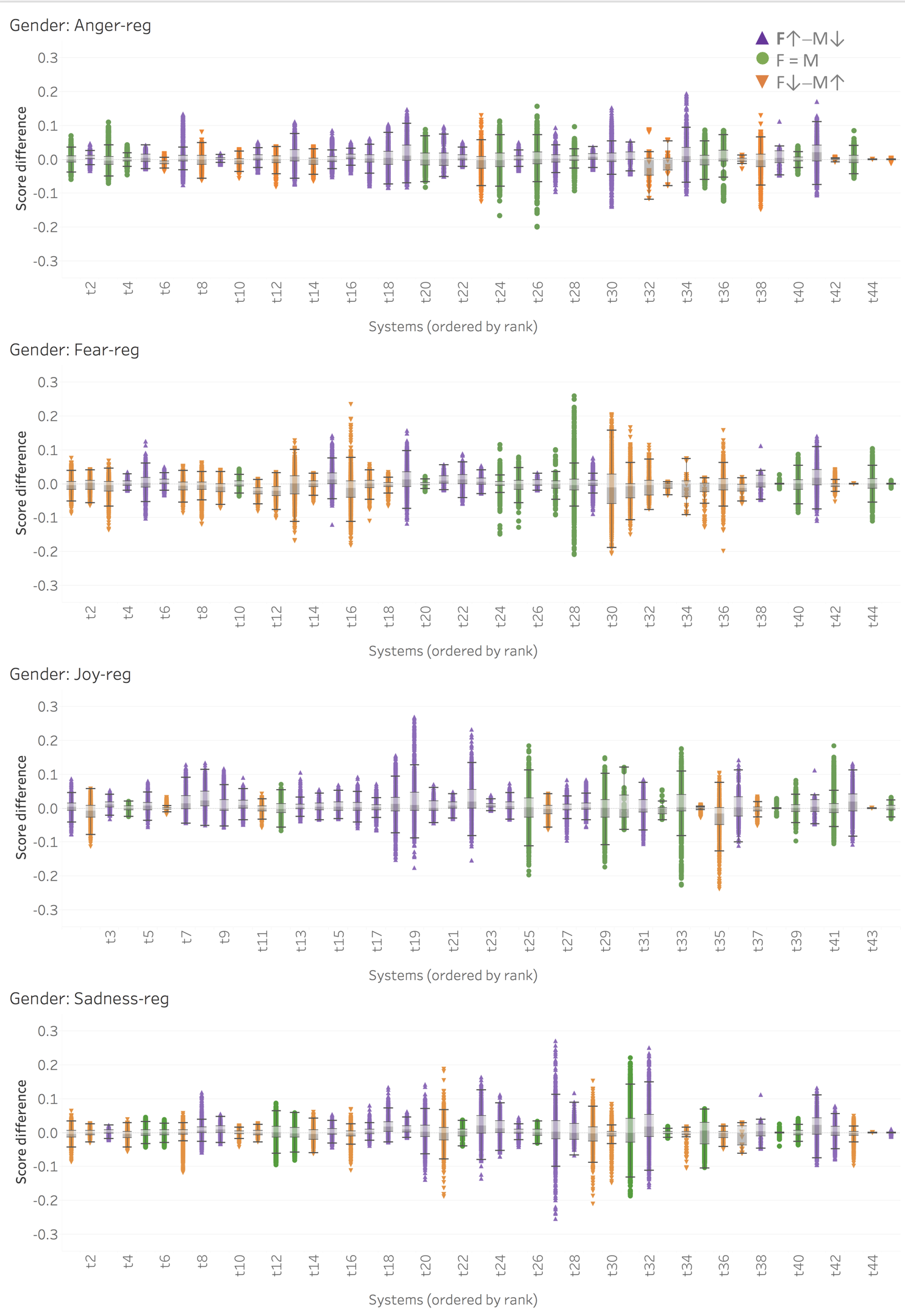}
\vspace*{-3mm}
\caption{\small \textbf{Analysis of gender bias:} Box plots of the score differences on the gender sentence pairs for each system on the four emotion intensity regression tasks. Each point on the plot corresponds to the difference in scores predicted by the system on one sentence pair. {\color{blue} \ding{115}} represents F$\uparrow$--M$\downarrow$ significant group, {\color{orange} \ding{116}} represents F$\downarrow$--M$\uparrow$ significant group, and {\color{green} \ding{108}} represents F=M not significant group. 
The systems are ordered by their performance rank (from first to last) on the task as per the official evaluation metric on the tweets test sets. 
The system with the lowest performance had the score differences covering the full range from -1 to 1, and is not included in these plots.
}
\vspace*{0mm}
\label{fig-score-diff-gender-EI}
\end{figure*}

\begin{figure*}[t]
\centering
\includegraphics[width=6in]{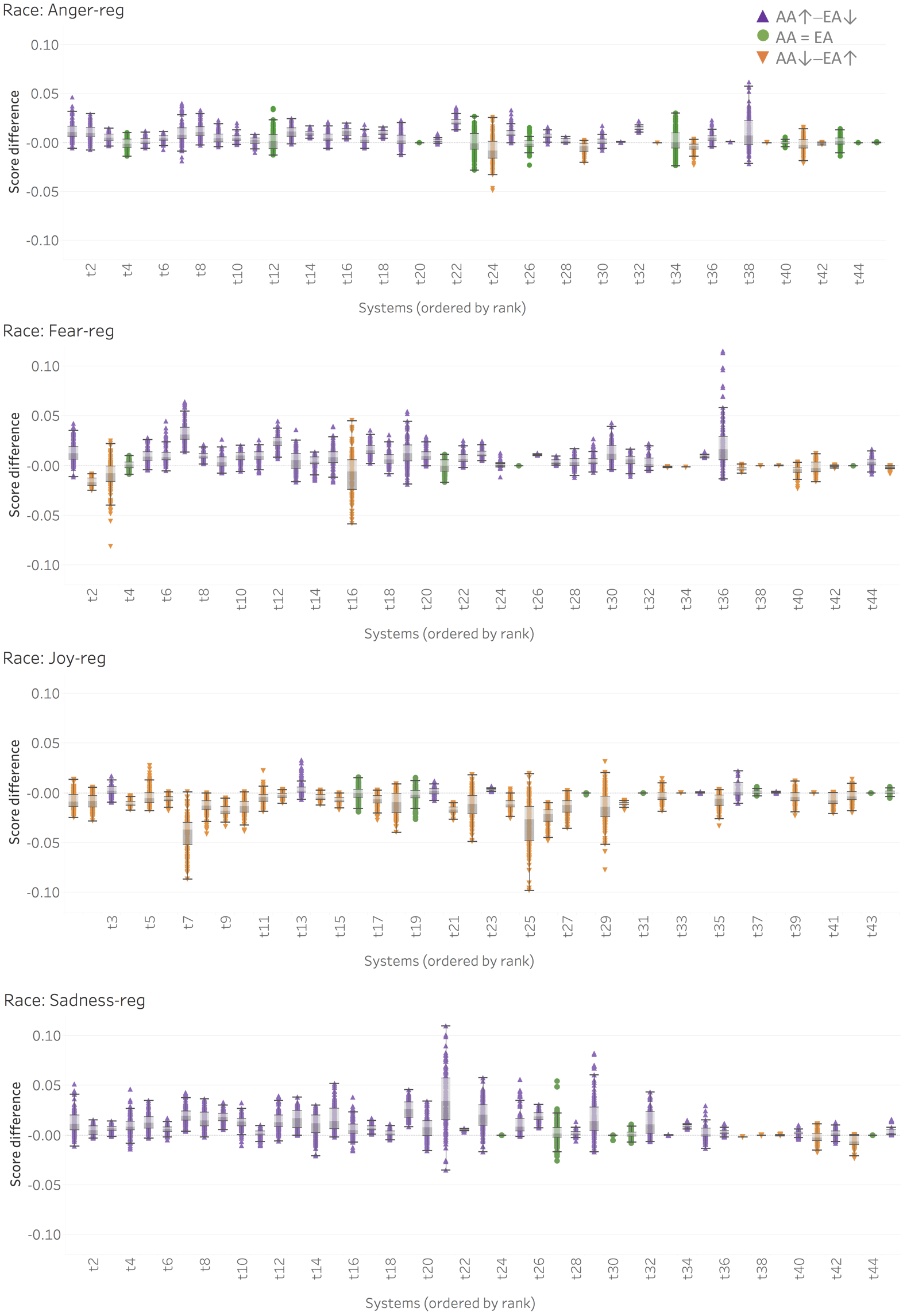}
\vspace*{-3mm}
\caption{\small \textbf{Analysis of race bias:} Box plots of the score differences on the race sentence pairs for each system on the four emotion intensity regression tasks. Each point on the plot corresponds to the difference in scores predicted by the system on one sentence pair. {\color{blue} \ding{115}} represents AA$\uparrow$--EA$\downarrow$ significant group, {\color{orange} \ding{116}} represents AA$\downarrow$--EA$\uparrow$ significant group, and {\color{green} \ding{108}} represents AA=EA not significant group. 
The systems are ordered by their performance rank (from first to last) on the task as per the official evaluation metric on the tweets test sets. 
The system with the lowest performance had the score differences covering a much larger range (from -0.3 to 0.3), and is not included in these plots.
}
\vspace*{0mm}
\label{fig-score-diff-race-EI}
\end{figure*}

Figures~\ref{fig-score-diff-gender-EI} and \ref{fig-score-diff-race-EI} show box plots of the score differences for each system on the four emotion intensity regression tasks on the gender and race sentence pairs, respectively. 
Each point on a plot corresponds to the difference in scores predicted by the system on one sentence pair. 
The systems are ordered by their performance rank (from first to last) on the task as per the official evaluation metric on the tweets test sets.

\end{document}